\documentclass{article}

\usepackage[preprint]{neurips_2022}

\usepackage[utf8]{inputenc} 
\usepackage[T1]{fontenc}    
\usepackage{hyperref}       
\usepackage{url}            
\usepackage{booktabs}       
\usepackage{amsfonts}       
\usepackage{nicefrac}       
\usepackage{microtype}      
\usepackage{xcolor}         

\usepackage{times}
\usepackage{soul}
\usepackage{url}
\usepackage{graphicx}
\usepackage{amsmath}
\usepackage{amsthm}
\usepackage{amssymb}
\usepackage{amsfonts}
\usepackage[ruled,vlined]{algorithm2e}
\usepackage[tableposition=top]{caption}
\usepackage{multirow}
\usepackage{booktabs}
\usepackage{enumitem}
\usepackage{footmisc}
\usepackage{subfigure}
\usepackage{appendix}
\usepackage{color}
\usepackage{multibib}
\usepackage{wrapfig}
\newcites{apndx}{References in Appendix}
\title{Task-Specific Expert Pruning for Sparse Mixture-of-Experts}

\author{
  Tianyu Chen$^{1,2}$, Shaohan Huang$^{4}$, Yuan Xie$^{3}$, Binxing Jiao$^{5}$, Daxin Jiang$^{5}$, \\
  \textbf{Haoyi Zhou$^{1,2}$, Jianxin Li$^{1,2}$, Furu Wei$^{4}$} \\
  $^{1}$BDBC, Beihang University, China \ $^{2}$SKLSDE Lab, Beihang University, China \\
  $^{3}$The Institute of Acoustics of the Chinese Academy of Sciences, China \\
  $^{4}$Microsoft Research \ $^{5}$NLP Group, Microsoft STCA 
}

\begin{document}

\maketitle

\begin{abstract}

The sparse Mixture-of-Experts (MoE) model is powerful for large-scale pre-training and has achieved promising results due to its model capacity. However, with trillions of parameters, MoE is hard to be deployed on cloud or mobile environment. The inference of MoE requires expert parallelism, which is not hardware-friendly and communication expensive. Especially for resource-limited downstream tasks, such sparse structure has to sacrifice a lot of computing efficiency for limited performance gains. In this work, we observe most experts contribute scarcely little to the MoE fine-tuning and inference. We further propose a general method to progressively drop the non-professional experts for the target downstream task, which preserves the benefits of MoE while reducing the MoE model into one single-expert dense model. Our experiments reveal that the fine-tuned single-expert model could preserve 99.3\% benefits from MoE across six different types of tasks while enjoying 2x inference speed with free communication cost. 

\end{abstract}

\section{Introduction}

\begin{figure}[!htbp]
    \centering
    \includegraphics[width=\linewidth]{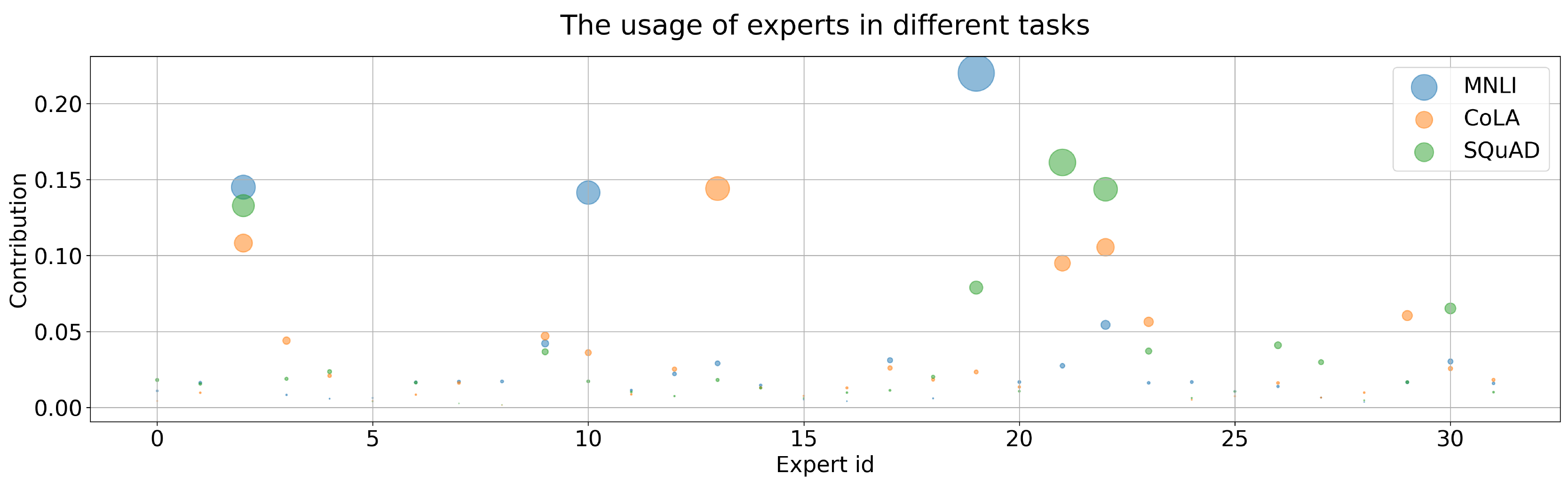}
    \caption{\textbf{The contribution of experts varies in different downstream tasks} The distribution of the contribution is quite long-tailed, indicating some experts are more "professional" on such tasks.}
    \label{fig:various_expert}
\end{figure}

In recent years, scaling up neural network models has brought significant quality improvements in both language~\cite{BERT,brown2020language} and vision domains~\cite{dosovitskiy2020image}. While large dense models have hit the boundary of model size and hardware capacity, another line of work has proposed sparsely-gated Mixture-of-Experts (MoE) layers as an efficient alternative to dense models~\cite{Gshards,switch_transformer,visual_moe}. In a vanilla sparsely-gated MoE model, each token of the input sequence activates a different subset of the experts, hence the computation cost per token becomes only proportional to the size of the activated sub-network. For a general pre-training task, the complexity and diversity of the optimization target require each sub-network (expert), to take a professional part of sub-task, thus increasing the model capacity and leading to generalization ability.

However, the proficiency of experts in MoE models is well under-explored in fine-tuning and serving stages. Different from a general and complex pre-training task, the downstream task for fine-tuning an MoE model is always specific and data-limited. Considering the diversity of experts in pre-trained MoE models, only a subset of experts will be well-activated, which may dominate the contribution to the input sequence. As depicted in Figure \ref{fig:various_expert}, even with the same pre-trained MoE model, the contribution of experts in MNLI, CoLA and SQuAD are quite different. Based on such observation, we propose an important question, \textit{For downstream tasks, could we select the most professional expert by certain fine-tuning strategies?}.

It may have several advantages to convert large sparse MoE models into small single-expert dense model by task-specific expert pruning method. First, the selected most professional expert for the task inherits the most transferable knowledge from MoE pre-training, which may contribute to better performance than the dense-training counterpart. Second, the single-expert model avoids the parallelism of experts, which decreases the requirements of devices and cut the cost of communication between devices, thus boosting the inference efficiency. Finally, no extra distillation effort should be paid, single-expert paradigm turned the MoE model into a dense counterpart architecture, compatible to a wide range of different downstream tasks.

In this work, we explore a general paradigm about how to find the most professional expert in MoE models during the fine-tuning stage, which aims to turn the sparsely-trained MoE model into a single-expert dense counterpart. We dive into the under-explored areas, including the criterion of pruning and the timing to drop experts. In an ideal scenario, we want to efficiently train a single large model maximizing the positive transfer while expanding the capacity bottleneck; meanwhile, we would like to enjoy the benefits of sparsely activated sub-networks per task at inference time. In practice, we set a dropping threshold at every $k$ steps of the fine-tuning process. If the proficiency score of one expert fails to meet the dropping threshold, the expert should be pruned. We repeat the dropping operation until there is only a single expert in every MoE layer or the fine-tuning process has passed half-way. Usually, the most professional expert of the task will have dominated the contribution before the end of  half fine-tuning process, thus we must drop all other non-professional experts and keep the selected expert for further optimization.

We summarize our main contributions as follows:
\begin{itemize}
    \item  We explore the task-specific expert pruning paradigm, a general device-friendly fine-tuning method for Mixture-of-Experts. We carefully drop the non-professional experts for the downstream task and keep the most professional one for inference and serving. We show that the selected expert could be positively transferred and successfully preserve most benefits from MoE pre-training.
    \item  We show that pruned MoEs outperform their dense counterparts on NLI, sentiment, similarity and QA tasks in absolute terms. Moreover, at inference time, the pruned MoEs can (i) match or even outperform the all-experts fine-tuned MoEs. and (ii) eliminate the communication cost between different devices.
    \item We provide visualization of the expert pruning during the fine-tuning process, revealing patterns and conclusions which helped motivate selection decisions and may further improve the final performance of fine-tuned models.
\end{itemize}

\section{Related Work}

\textbf{Sparse Mixture-of-Experts Models} Sparse Mixture-of-Experts (MoE) models that provide a highly efficient way to scale neural network training, have been studied recently in~\cite{Gshards,switch_transformer,shazeer2017outrageously}.
Compared to standard dense models that require extremely high computational cost in training, MoE models are shown to be able to deal with a much larger scale of data and converge significantly faster. Some recent studies focus on the expert assignment problem, such as formulating token-to-expert allocation as a linear assignment~\cite{lewis_base_nodate}, hashing-based routing~\cite{roller_hash_nodate}, and distillation in routing function~\cite{dai2022stablemoe}. In addition, a few recent works show different training methods for sparse MoE models. \citet{dua_tricks_2021} found that a temperature heating mechanism and dense pre-training can improve the performance of sparse translation models. \citet{nie_dense--sparse_2021} propose a dense-to-sparse gating strategy for MoE training. Some papers have explored the performance of supervised fine-tuning in MoE models and proposed some methods to boost the quality of fine-tuning. \citet{artetxe_efficient_nodate} introduced an expert regularization to improve sparse model fine-tuning. \citet{chi2022representation} proposed a dimension reduction and $L_2$ normalization to solve the representation collapse in fine-tuning. Instead of improving fine-tuning of MoE models, this work focuses on fine-tuning spare models into single-expert dense models directly, namely task-specific expert pruning .


\textbf{Efficient Inference in MoE Models}  MoE models have achieved promising results. However, they fail to meet requirements on inference efficiency in fine-tuning and serving stage. Recent studies apply knowledge distillation for faster inference. \citet{rajbhandari_deepspeed-moe_nodate} proposed a smaller MoE model (named Mixture-of-Students) with knowledge distillation. However, sparse MoE makes inference difficult for the following reasons: if the mode parameters exceed the memory capacity of a single accelerator device, inference requires multiple devices.
Communication costs across devices further increase the overall serving cost. There have been some works that apply distillation of sparse MoE models into dense models~\cite{fedus_switch_2021,xue_one_nodate,artetxe_efficient_nodate}. \citet{fedus_switch_2021} presents a study of distilling a fine-tuned sparse model into a dense model. The task-specific distillation method needs to fine-tune different models for different tasks respectively. This work does not require additional fine-tuning on MoE models and adapts sparse MoE models into single-expert dense models directly.


\section{Preliminary}
\label{sec:preliminary}

Recall the Mixture-of-Expert Transformer models proposed by Shazeer et al.~\cite{shazeer2017outrageously}, where the MoE layer takes an input token representation $x$ and then routes this to the best determined top-$k$ experts selected from a set $\{Exp_i(x)\}_{i=1}^E$ of $E$ experts. The logits $h(x) = W_r \cdot x$ produced by the router variable $W_r$ is normalized via a softmax distribution over the available $E$ experts at that layer. The gated-value of expert $i$ is given by,

\begin{equation}
    \alpha_i(x) = \frac{e^{h(x)_i}}{\sum_j^E{e^{h(x)_j}}}
\end{equation}

The top-$k$ gate values are selected for routing the token $x$. if $\mathcal{P}$ is the set of selected top-$k$ indices then the output computation of the layer is the linearly weighted combination of each expert's computation on the token by the gated value, 

\begin{equation}
    y  = \sum_{i \in \mathcal{P}}{\alpha_i(x)Exp_i(x)}
\end{equation}

\textbf{Notation} \ In this paper, the symbol $E$ denotes the total number of experts while $e$ denotes the id of the expert in one Mixture-of-Experts layer.  $N$ denotes the total training steps and $w$ denotes one specific training window. $Z$ denotes the number of survival experts in the current training window. 

\textbf{Problem Formulation} \ Given the labeled data of one downstream task and a pre-trained MoE model with several mixture-of-experts layers, where each MoE layer has $E$ experts, we try to find the most professional expert $\hat{e}$ for the task in every MoE layer and prune the other non-professional experts, during fine-tuning the MoE model on the labeled data.

\section{Method}
\label{sec:Method}

\subsection{Overview}

\begin{figure}
    \centering
    \includegraphics[width=\linewidth]{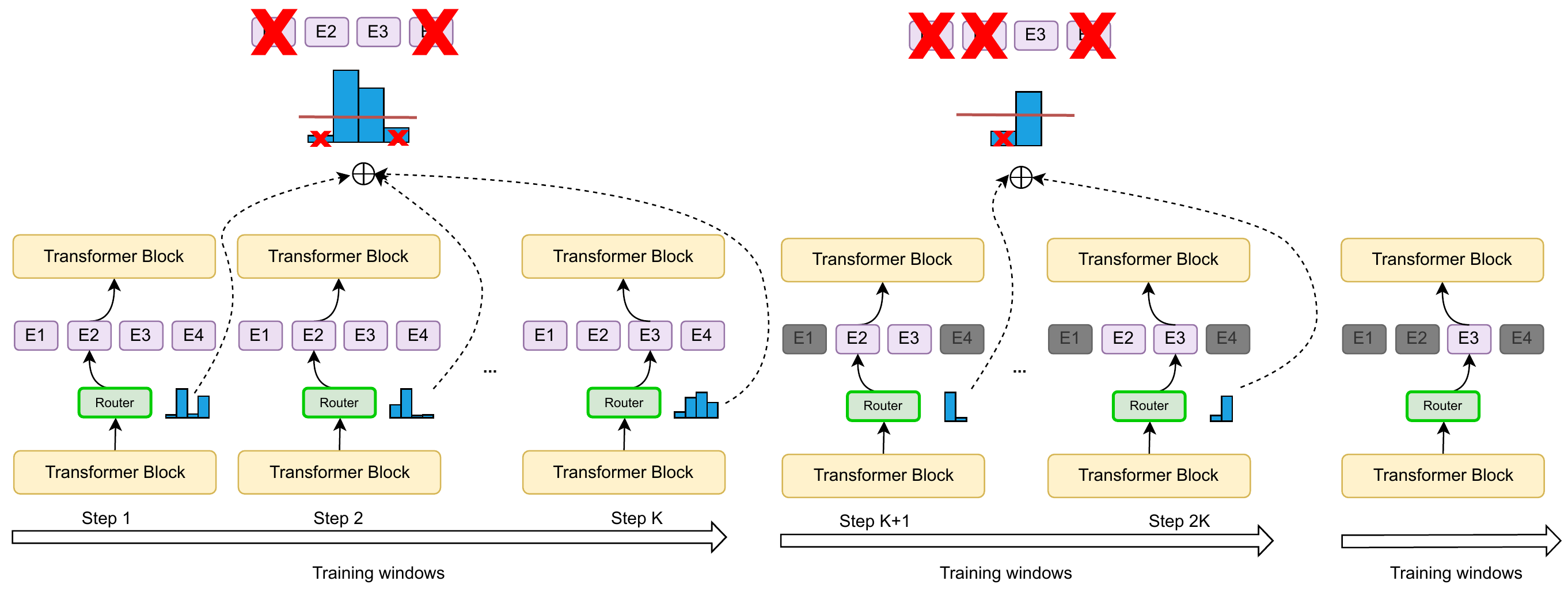}
    \caption{The pipeline of our method. We split the fine-tuning process into several  training windows evenly and prune the non-professional experts at the end of each training window. }
    \label{fig:pipeline}
\end{figure}

We illustrate our method in Figure~\ref{fig:pipeline}, where we do not break the the standard pipeline of fine-tuning an MoE model on the downstream tasks. Instead, we split the $N$ training steps of the whole fine-tuning process into $E$ training windows evenly.  During one training window, we will accumulate the proficiency score of each expert when the representations of the input tokens pass through the MoE layer. At the end of one training window, we decide which experts are non-professional and should be pruned. The pruned experts will never participate in the assignment of tokens at the later training windows, also will be dropped for inference. At the half point of the whole training schedule, we make the final decision and keep only the most professional expert in each MoE layer and prune all other experts. Once the most professional expert is selected, all input tokens are assigned to the expert and reduce the sparse MoE model into a dense counterpart.


\subsection{The Criterion of Expert Proficiency}
\label{sec:criterion}

The first critical point of our method is the evaluation of expert proficiency. Note that when input with token sequence $x_s$, the MoE model will execute two operations related to the schedule of experts. The first operation is the assignment of tokens, where each token can only be assigned to a single expert. The second operation is conquering the division results, which uses the $\alpha_i$ to weighted sum the selected expert output and formulates the final output $y_s$.  During fine-tuning of MoE models, both of the expert schedule-related operations are criterion of expert proficiency. We define them respectively as hit rate $R$ and alpha score $\mathcal{A}$.

 In our method, we choose $\mathcal{A}$ as the default criterion and define the proficiency score of expert $i$ in one training window as $C_{i}$:

\begin{equation}
    C_{i} = \sum_{k}^{k+N/E} \mathcal{A}_{i}
\end{equation}


\begin{equation}
    \mathcal{A}_{i} = \frac{\alpha_i}{\sum_{i}^{E}{\alpha_i}}
\end{equation}

Here, $k, N$ represent the start step of the training window and the total training steps in the fine-tuning schedule respectively. Both $R_i$ and $\mathcal{A}_i$ are accumulated during a training window of $K$ training steps. As more tokens are assigned to one expert $i$ and the gating alpha score $\mathcal{A}_i$ gets higher,  the expert $i$ is more critical to the final output $y$, thus leading the prediction on the downstream task.  A higher hit rate $R_i$ will yield more input tokens to be assigned to the expert $i$ and fewer tokens to be computed by other experts, also may give proper approximation of the expert proficiency on the downstream tasks.

\subsection{The Timing to Drop Non-Professional Experts}

To find the most professional expert, one intuitive method is two-pass training, which selects the expert in the first pass and fine-tunes the dense model in the second pass. However, different from post-training pruning or distillation strategy, our method is pursuing a one-pass fine-tuning paradigm, which does not incur extra computation after fine-tuning an MoE model. In our during-training strategy, we find when to drop the non-professional experts greatly affects the final performance of fine-tuned model. In one respect, if we accumulate the proficiency score $\mathcal{C}$ in a relatively long training window, the approximation of the expert proficiency will be more reliable and closer to the whole task. In another respect, the early decision to drop non-professional experts leaves plenty of training room for optimizing the chosen professional expert and avoid wasting computing on non-professional experts. Thus in our one-pass strategy,  the timing to drop is a trade-off between the more appropriate selection of the professional expert and the more optimization of the final expert.

In practice, we propose combining the progressive dropping and the force dropping. We divide the whole training steps evenly into $E$ windows and progressively make the dropping decision at the end of each window. If $E/2$ windows have passed and the MoE layer still has more than one survival experts, only the expert with the highest proficiency score will be kept.

\subsection{The Progressively Dropping}

During progressive dropping, we have $E/2$ chances to make the dropping decision.  It is an important problem in each dropping decision that how many experts should be dropped. To select out the most non-professional experts in each training window, we propose a dynamic dropping threshold $\mathcal{T}$. Any expert with a proficiency score below the threshold $\mathcal{T}$ will be dropped.

\begin{equation}
    \mathcal{T} = \beta * 1/Z
\end{equation}

Here, $Z$ denotes the number of survival experts in the current training window. $\beta$ is a hyper-parameter to control the speed of the progressively dropping. A larger $\beta$ will increase the threshold $\mathcal{T}$ and more experts will be dropped in each decision. In our experiments, a relatively small $\beta$ maybe a good default to preserve maximum performance of MoE models.

\section{Experiments}

In this section, we design both sequence-level and token-level tasks to evaluate the performance and efficiency of our task-specific expert pruning paradigm. We also provide a detailed discussion about the hyper-parameters that affect our strategy.

\subsection{Evaluation Tasks}
\label{sec:evaluation_tasks}

\textbf{GLUE}~\cite{GLUE}, the General Language Understanding Evaluation benchmark, is a collection of tools for evaluating the performance of models across a diverse set of existing NLU tasks, including STS-B~\cite{STS-B}, CoLA~\cite{CoLA}, MRPC~\cite{MRPC}, RTE~\cite{RTE1, RTE2, RTE3, RTE4}, SST-2~\cite{SST2}, QQP, MNLI~\cite{MNLI} and QNLI~\cite{SQUAD1.1}.  Following previous works~\cite{BERT, turc2019well}, we exclude the WNLI task from the GLUE benchmark. Considering the small data size of RTE may lead to significant bias in performance evaluation, we exclude RTE results in Table \ref{tab:main_results}, but include all GLUE tasks in Appendix Table~\ref{tab:more_results}.

\textbf{SQuAD2.0}~\cite{SQUAD2.0}, the Stanford Question Answering Datasets, is a collection of question-answer pairs derived from Wikipedia articles. In SQuAD, the correct answers can be any sequence of tokens in the given tasks. SQuAD2.0, the latest version, combines the 100,000  questions in SQuAD1.1 with over 50,000 unanswerable questions written by crowd workers in forms that are similar to the answerable ones.

\subsection{Experiments Settings}

We use the base Switch Transformer~\cite{switch_transformer} architecture for this set of experiments. For a fair comparison, we also employ a dense transformer model of the same architecture as Switch-base except for the mixture-of-experts layers. We pre-train both the MoE and the dense model for 125k steps on the BooksCorpus~\cite{Zhu_2015_ICCV} and English Wikipedia corpus~\cite{wikidump}, with the masked language modeling objective. For the MoE model, we use a MoE layer of 32 experts in the 4th and 8th transformer block. A more detailed pre-training settings and model hyperparameters could be found in Appendix~\ref{app:params:pt}.

Here we include several fine-tuning settings for comparison. All fine-tuning hyperparameters can be found in~Appendix~\ref{app:params:ft}:

\textbf{Dense-ft:} The standard dense model fine-tuning, which is for the demonstration of MoE benefits and as a strong baseline for our expert pruning strategy.

\textbf{MoE-ft:} The standard MoE fine-tuning method, where all 32 experts participate in the optimization for the downstream task and also for the final inference.

\textbf{Two-pass-staged-drop:} An intuitive two-pass pruning baseline, where we first accumulate the proficiency score of each expert in the first pass fine-tuning and keep the most professional expert to fine-tune the MoE model again.

\textbf{Two-pass-eager-drop:} A baseline to prove the effectiveness of our proficiency criterion, where we select the most professional expert based on our pipeline and keep the selected expert to re-fine-tune the MoE model.

\textbf{MoE-staged-pruning:} A task-specific expert progressively pruning strategy, where we drop one most non-professional expert at the end of each training window, yields a single-expert model for inference.

\textbf{MoE-eager-pruning:} The method implements all parts of our task-specific expert pruning pipeline, which will drop the experts of proficiency score below the bar at each training window and execute a force drop at the latter half training schedule.

\subsection{Preserving Benefits of MoE into a Single Expert}

\begin{table}[htbp]
  \centering
  \caption{\textbf{The performance on the GLUE tasks for the dense models and the MoE models}, measured on the development sets. We report the average results by a set of seeds. \textbf{AVG} denotes the average score of  all the tasks. Task-specific expert pruned MoE models with only single expert could achieve about 0.84 avg glue score than dense models. }
  \resizebox{\linewidth}{!}{
   \begin{tabular}{cccccccccc}
    \toprule
    \textbf{Settings} & \textbf{CoLA} & \textbf{STS-B} & \textbf{MNLI-m} & \textbf{MNLI-mm} & \textbf{SST-2} & \textbf{QQP} & \textbf{QNLI} & \textbf{MRPC} & \textbf{AVG } \\
    \midrule
    Dense-ft & 57.23  & 89.18  & 85.87  & 85.87  & 92.87  & 91.20  & 92.23  & 88.07  & 85.32  \\
    MoE-ft & 63.75  & 89.37  & 86.93  & 86.93  & 94.03  & 91.37  & 92.90  & 89.13  & 86.80  \\
    \midrule
    Two-pass-staged-drop & 58.32  & 88.64  & 86.30  & 85.90  & 90.00  & 90.50  & 91.40  & 86.90  & 84.75  \\
    Two-pass-eager-drop & 60.89  & 89.02  & 86.10  & 86.20  & 92.60  & 90.60  & 91.50  & 87.72  & 85.58  \\
    \midrule
    MoE-staged-pruning & 56.20  & 89.00  & 85.33  & 85.33  & 92.33  & 90.07  & 91.97  & 86.83  & 84.63  \\
    \textbf{MoE-eager-pruning} & \textbf{61.47} & \textbf{89.43} & \textbf{86.10} & \textbf{86.10} & \textbf{93.17} & \textbf{91.20} & \textbf{92.53} & \textbf{89.30} & \textbf{86.16} \\
    \bottomrule
    \end{tabular}}
  \label{tab:main_results}
\end{table}

In Table \ref{tab:main_results}, the results can be divided into three groups. The first group is the baseline, where both sparsely pre-trained MoE models and their densely pre-trained counterparts are fine-tuned in standard ways. We observe the MoE enjoys full advantage at all tasks, of an 1.48 average glue score over their densely pre-trained counterparts, which may thank to the strong generalization ability of mixture-of-experts pre-training.

In the second group, we compare two intuitive expert pruning strategies based on a two-pass optimization. With more computation budget of another pass, we could select the most professional expert in the first pass while optimizing the selected expert in the second pass.  We find for the two pas staged drop method, another pass could boost the final performance of the model by 0.12 average GLUE score. For two pass eager drop, another pass hurts the performance, cutting down the average GLUE score by 0.58. We assume another pass gives plenty of room for optimization of the final selected expert in the staged drop, which maybe sub-optimal yet in the first pass. However, for the two pass eager drop, the selected expert is may well optimized and tend to overfit in the second training pass.

In the last group, we are interested in whether our method could preserve most advantage of mixture-of-experts by keeping only single most professional expert. We find in the MoE staged pruning setting, the selected expert fails to preserve the advantage of the MoE and only achieves 84.63 average score, 0.69 below the densely pre-trained counterpart. However, for the MoE eager pruning, the selected expert could preserve most benefits from MoE pre-training and outperforms the densely pre-trained models by 0.84 average GLUE scores. We assume the failure of the staged strategy may be due to the lack of enough adaptation of the final selected expert, which we will further discuss in section \ref{sec:timing}. In addition, we find the single expert selected by the eager method could even exceed the performance of standard fine-tuned MoE models, which have 32 experts, in STS-B and MRPC tasks. Considering the excessive model capacity of the MoE models, fine-tuning with all experts may not always be a good default for the data-limited downstream tasks.


\begin{table}[htbp]
  \centering
 \caption{The performance on SQuAD2.0 task. The EM denotes the accuracy of exactly matched answers.}
 
    \begin{tabular}{ccc}
    \toprule
    \textbf{Settings} & \textbf{SQuAD2.0-F1} & \textbf{SQuAD2.0-EM} \\
    \midrule
    MoE-ft & 80.59 & 77.85 \\
    Dense-ft & 79.77 & 77.23 \\
    \midrule
    Two-pass-drop & 80.11 & 77.31 \\
    Two-pass-eager-drop & 80.33 & 77.40 \\
    \midrule
    MoE-staged-pruning & 80.08 & 77.24 \\
    \textbf{MoE-eager-pruning} & \textbf{80.40} & \textbf{77.55} \\
    \bottomrule
    \end{tabular}
    \label{tab:squad_results}
\end{table}

For the token-level task, we present our results of SQuAD 2.0 in Table \ref{tab:squad_results}. We find the sparsely pre-trained MoE models exceed their densely counterparts by 0.82 F1 score and  0.62 EM score. For our task specific expert pruning method, both staged and eager pruning strategy exhibits comparable or better performance than the dense-ft baseline. The results of the SQuAD 2.0 is consistent with those in sentence-level GLUE benchmark, where the eager pruning outperforms the staged pruning by 0.32 F1 score and 0.31 EM score. In addition, the two pass optimization also outperforms the dense-ft baseline, where the two-pass-eager-drop exceeds dense-ft by 0.56 F1 score and 0.17 EM score. We assume token-level tasks are more expert-sensitive and our expert pruning method easily locates the most professional expert, leading to better performance. 


\subsection{Different Pruning Criterion}
\label{sec:selection methods}

\begin{figure}
    \centering
    \includegraphics[width=\linewidth]{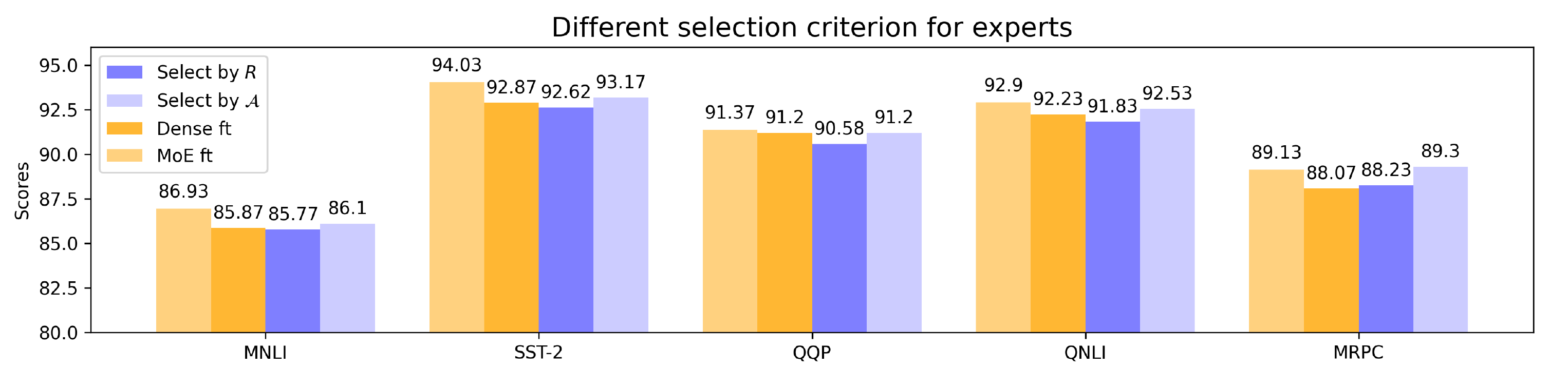}
    \caption{\textbf{Different pruning criterion for professional experts.} We compare the most professional expert chosen by hit rate $R$ and alpha score $\mathcal{A}$ on five different tasks. The most professional expert chosen by $\mathcal{A}$ better narrow preserve the performance of standard MoE-ft.}
    \label{fig:selection_criterion}
\end{figure}

As mentioned in Section \ref{sec:criterion}, both hit rate $R$ and alpha score $\mathcal{A}$ could be the evaluation of expert proficiency.  We present the results of these two different pruning criterion in Figure \ref{fig:selection_criterion}.  The $R$ based pruning only exceeds the dense-ft baseline on the MRPC task by 0.17 accuracy score while performing worse in the other five tasks. However, the $\mathcal{A}$ based pruning shows a much better performance across all five tasks and achieves better performance than the MoE-ft baseline on MRPC. 

We assume more tokens assigned to one expert may not well represent its contribution because the output of the expert computation on token $x$ will be re-weighted by gating score $\alpha_i(x)$. Thus the $\mathcal{A}$ could be viewed as "soft" $R$ with a more precise evaluation of the expert proficiency.

\subsection{A Closer Look at the Dropping Timing}
\label{sec:timing}

\begin{figure}
    \centering
    \includegraphics[width=0.6\linewidth]{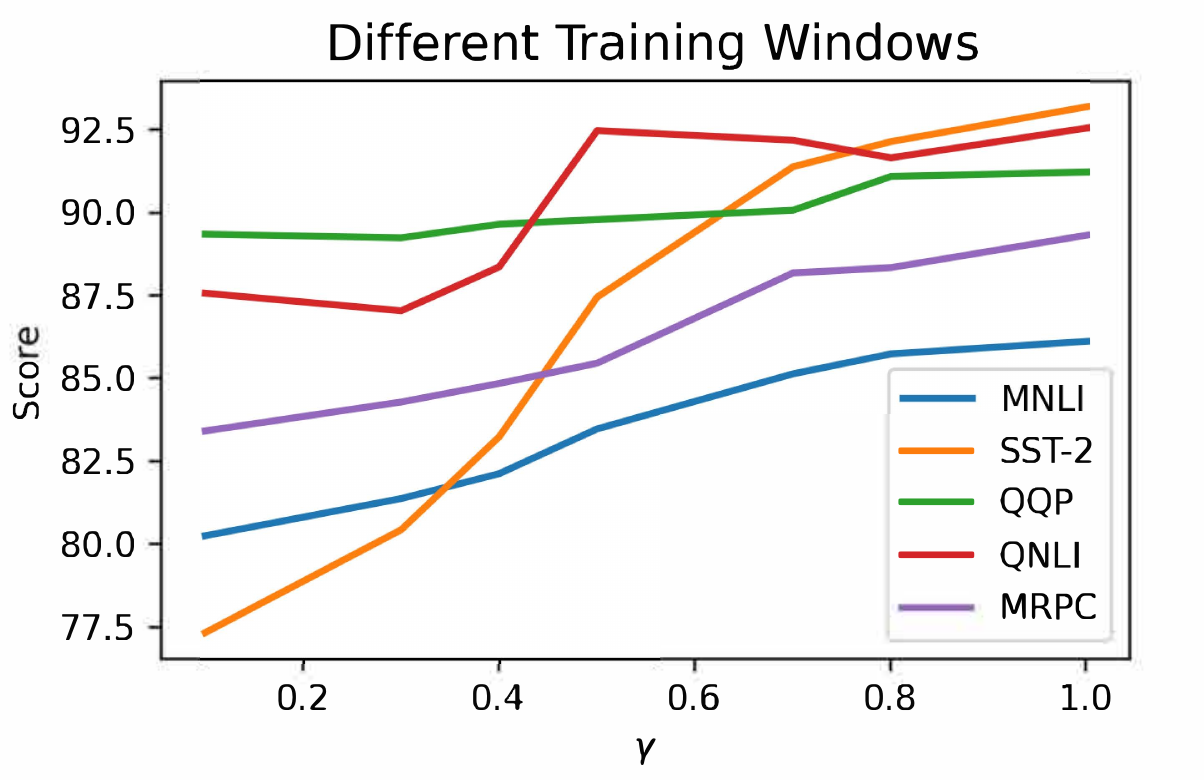}
    \caption{\textbf{The effect of training window length on GLUE tasks.} $\gamma$ is a parameter to control the length of training window $\mathcal{L}$, where $\mathcal{L} = \gamma * \frac{N}{E}$. $N$ and $E$ denote the total training steps and the number of experts respectively.}
    \label{fig:interval_gamma}
\end{figure}


One important question related to task-specific expert pruning is \textit{when should we make the final decision to drop all other experts.} Recall that in our methods, two factors could affect the final decision. The first factor is the length of the training window and the second factor is the dropping threshold.

As depicted in Figure \ref{fig:interval_gamma}, we investigate how long the training windows should be split to achieve better final performance on MNLI, SST-2, QQP, QNLI and MRPC tasks. We observe that for MNLI, SST-2, QQP and MRPC tasks, a longer training window contributes to better final performance of selected expert. While for the QNLI task, a training window length of $0.5 * \frac{N}{E}$ could also achieve comparable performance of a maximum training window. We assume for most tasks, longer training windows are beneficial for boosting the performance of the selected expert.

\begin{figure}[!htbp]
    \centering
    \includegraphics[width=\linewidth]{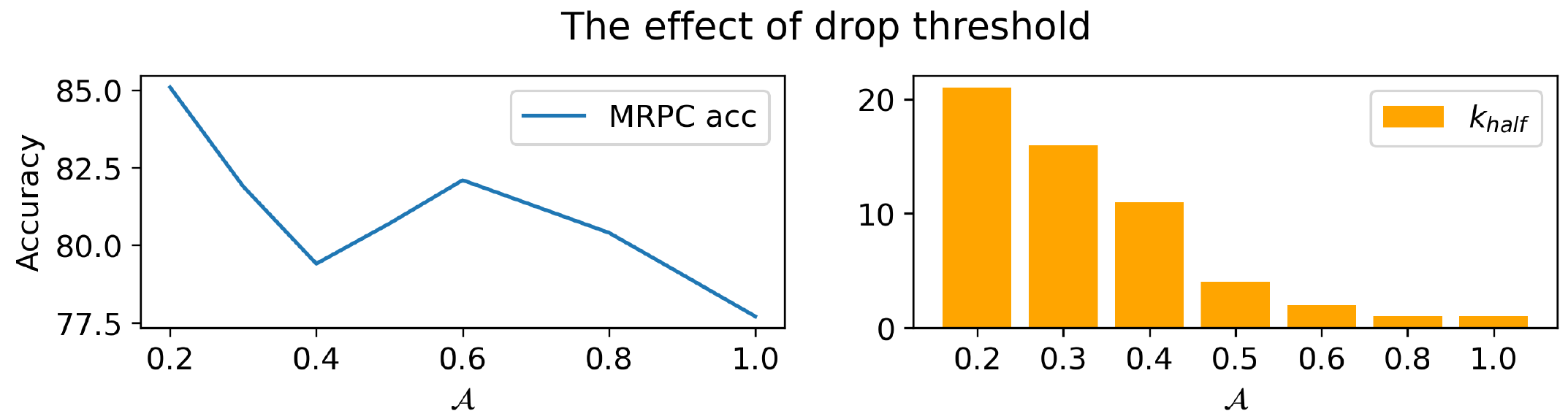}
    \caption{\textbf{Comparison of different drop threshold.} On the left figure, we illustrate the fine-tuned MoE accuracy on MRPC task with a different $\mathcal{A}$ threshold. On the right figure, we depict the number of survival experts at half of the schedule, which denoted as $K_{half}$.}
    \label{fig:alpha_selection}
\end{figure}

Another way to control the progress of the pruning is the dropping threshold. We show the effect of different drop thresholds in Figure \ref{fig:alpha_selection}. A lower dropping threshold may lead to slow progress of expert selection and more survival experts to be forced to drop at the half training schedule, but provides  more precise observation of experts contribution.

\begin{figure}[!htbp]
    \centering
    \includegraphics[width=\linewidth]{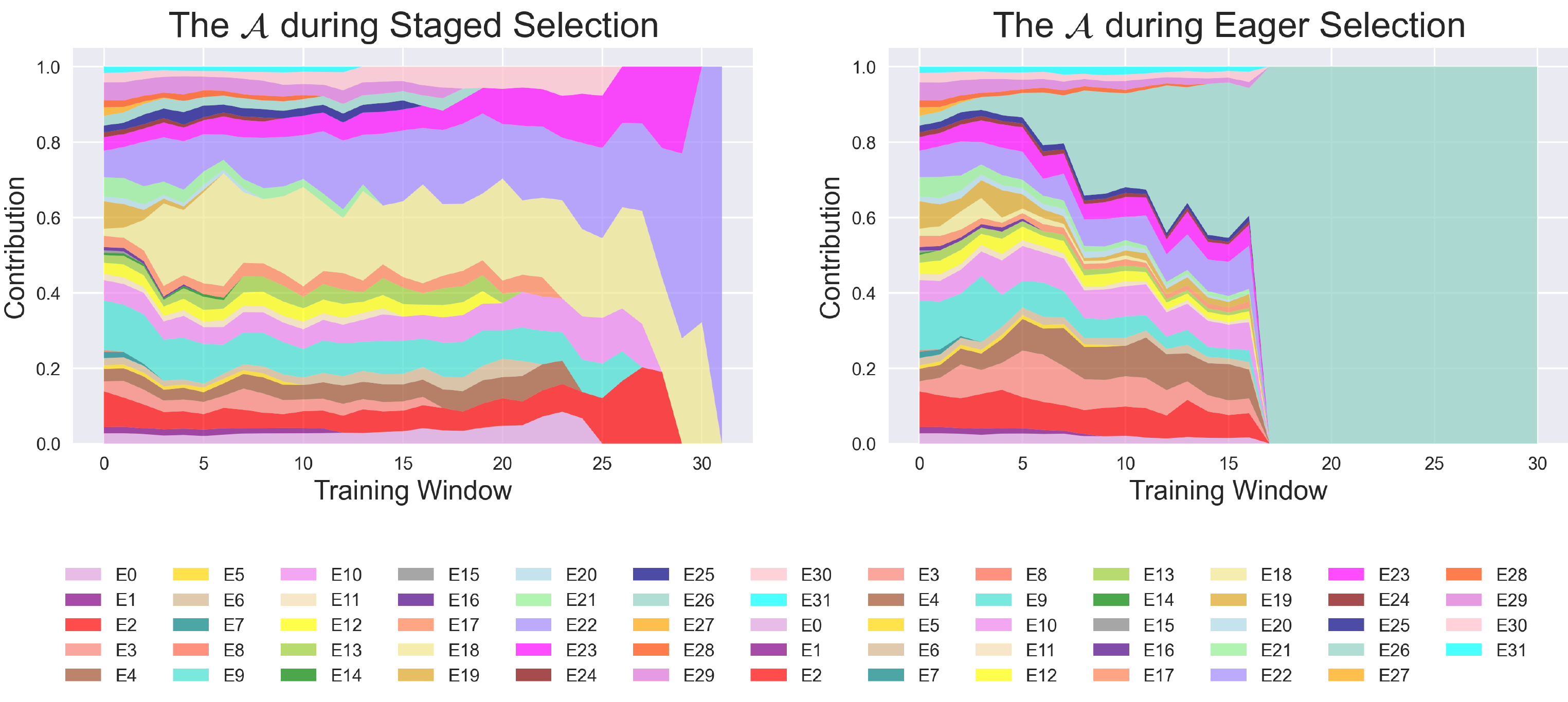}
    \caption{\textbf{Comparison of the timing for making final expert decision}. In the left sub-figure, we illustrate the staged pruning, which adopts an "lazy" pruning strategy and make the final decision at the last training window. In the right sub-figure, we illustrate a more "eager" strategy, which drops more experts in the first several training windows and leave the last half training schedule for the most professional expert.}
    \label{fig:alpha_distribution}
\end{figure}

Finally, we compare the timing for making the final expert decision in Figure~\ref{fig:alpha_distribution}. For the staged pruning method, we find some experts gradually dominate the contribution to the final output during training. However, even if the final selected expert has dominate the contribution in early training window, it could not be optimized independently until the last training window, which may answer the relatively poor performance in inference stage. Based on such observation, we propose a more eager pruning method, which drops all non-professional experts at the half of training since our selected expert is dominant in the contribution. This method leaves the later half of the training schedule to optimize the selected expert, which leads to a better inference performance than the staged method.

\subsection{The Efficiency of Inference}
\label{sec:inference}
\begin{wrapfigure}{R}{0.5\textwidth}
    \begin{center}
        \includegraphics[width=0.5\textwidth]{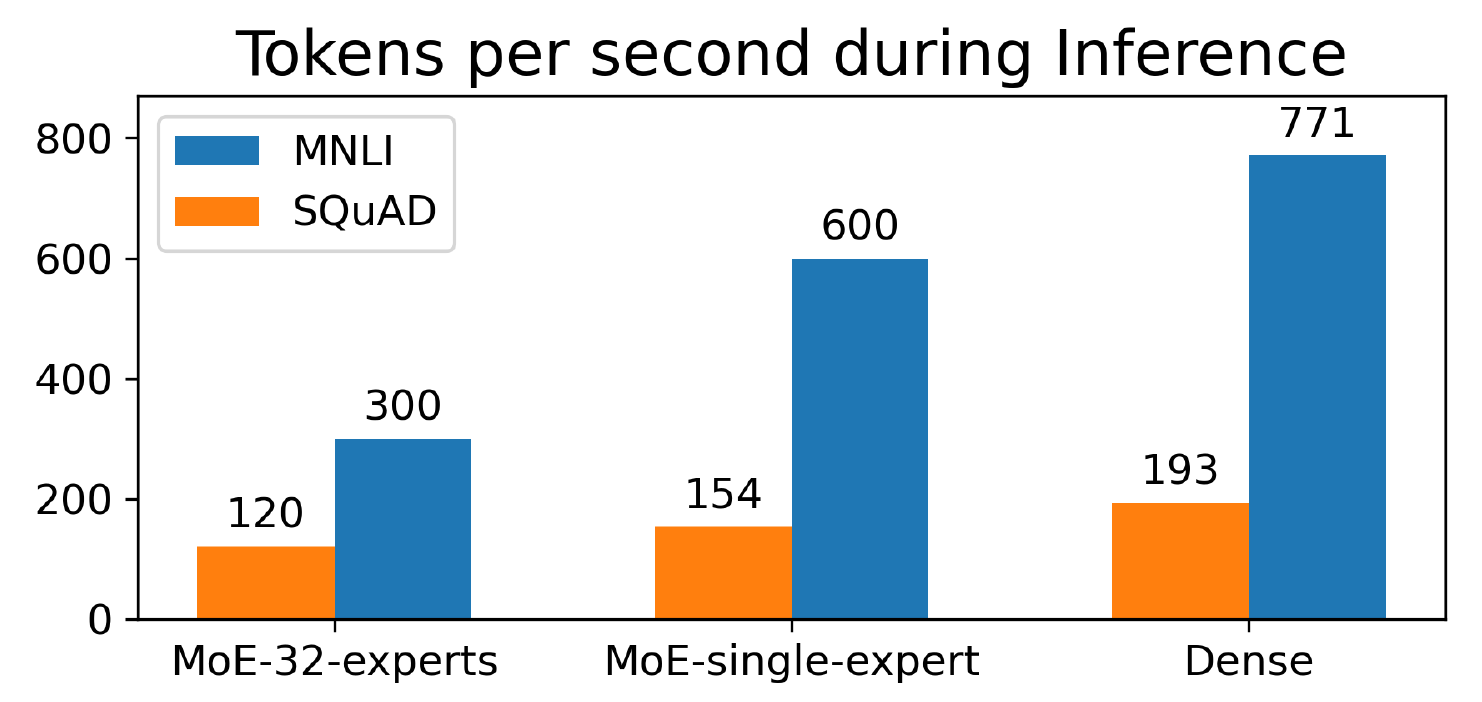}
        \caption{The inference speed of different fine-tuned models.}
        \label{fig:inference}
    \end{center}
\end{wrapfigure}

We present the inference speed of different settings in Figure~\ref{fig:inference}, evaluated on MNLI and SQuAD tasks. We use the tokens per second as an metric of the model inference speed and use NVIDIA-V100-32GB for testing. For the MoE model fine-tuned with 32 experts, the extra communication cost between different devices greatly reduces the inference speed, leading to the worst inference efficiency among all settings. The single expert models optimized with our task-specific expert pruning method, enjoy an inference efficiency advantage of 2x speed up on MNLI task and 1.28x speed up on SQuAD task. Because of the extra sub-MoE layers in the selected expert, our method slightly lags behind the inference efficiency of the dense model, achieving 80\% inference speed of dense model on MNLI and SQuAD tasks.

\section{Conclusion}

In this paper, we discussed a more inference friendly fine-tuning paradigm for the pre-trained Mixture-of-experts models by fine-tuning the most professional expert and drop other experts, namely task-specific expert pruning. We empirically demonstrate that this new fine-tuning paradigm could preserve most benefits of the pre-trained MoE models and much better than the densely pre-trained counterparts, on both sentence-level tasks and token-level tasks. By carefully examining the factors of expert pruning across different settings, we demonstrate the superiority of our eager expert pruning paradigm over other possible solutions like two-pass optimization or staged expert pruning.

We conclude by highlighting the task specific expert pruning paradigms that are more inference friendly while retaining the quality gains of MoE models are promising direction of future exploration.

\clearpage




\bibliographystyle{plainnat}
\small
\bibliography{main}

\begin{thebibliography}{32}
\providecommand{\natexlab}[1]{#1}
\providecommand{\url}[1]{\texttt{#1}}
\expandafter\ifx\csname urlstyle\endcsname\relax
  \providecommand{\doi}[1]{doi: #1}\else
  \providecommand{\doi}{doi: \begingroup \urlstyle{rm}\Url}\fi

\bibitem[Artetxe et~al.()Artetxe, Bhosale, Goyal, Mihaylov, Ott, Shleifer,
  Victoria~Lin, Du, Iyer, Pasunuru, Anantharaman, Li, Chen, Akin, Baines,
  Martin, Zhou, Singh~Koura, Wang, Zettlemoyer, Diab, Kozareva, and
  Stoyanov~Meta]{artetxe_efficient_nodate}
Mikel Artetxe, Shruti Bhosale, Naman Goyal, Todor Mihaylov, Myle Ott, Sam
  Shleifer, Xi~Victoria~Lin, Jingfei Du, Srinivasan Iyer, Ramakanth Pasunuru,
  Giri Anantharaman, Xian Li, Shuohui Chen, Halil Akin, Mandeep Baines, Louis
  Martin, Xing Zhou, Punit Singh~Koura, Jeff Wang, Luke Zettlemoyer, Mona Diab,
  Zornitsa Kozareva, and Ves~AI Stoyanov~Meta.
\newblock Efficient {Large} {Scale} {Language} {Modeling} with {Mixtures} of
  {Experts}.
\newblock URL \url{https://github.com/pytorch/fairseq/}.
\newblock arXiv: 2112.10684v1 ISBN: 1.302410245.

\bibitem[Bentivogli et~al.(2009)Bentivogli, Magnini, Dagan, Dang, and
  Giampiccolo]{RTE4}
Luisa Bentivogli, Bernardo Magnini, Ido Dagan, Hoa~Trang Dang, and Danilo
  Giampiccolo.
\newblock The fifth {PASCAL} recognizing textual entailment challenge.
\newblock In \emph{{TAC}}. {NIST}, 2009.

\bibitem[Brown et~al.(2020)Brown, Mann, Ryder, Subbiah, Kaplan, Dhariwal,
  Neelakantan, Shyam, Sastry, Askell, et~al.]{brown2020language}
Tom Brown, Benjamin Mann, Nick Ryder, Melanie Subbiah, Jared~D Kaplan, Prafulla
  Dhariwal, Arvind Neelakantan, Pranav Shyam, Girish Sastry, Amanda Askell,
  et~al.
\newblock Language models are few-shot learners.
\newblock \emph{Advances in neural information processing systems},
  33:\penalty0 1877--1901, 2020.

\bibitem[Cer et~al.(2017)Cer, Diab, Agirre, Lopez{-}Gazpio, and Specia]{STS-B}
Daniel~M. Cer, Mona~T. Diab, Eneko Agirre, I{\~{n}}igo Lopez{-}Gazpio, and
  Lucia Specia.
\newblock Semeval-2017 task 1: Semantic textual similarity multilingual and
  crosslingual focused evaluation.
\newblock In \emph{SemEval@ACL}, pages 1--14. Association for Computational
  Linguistics, 2017.

\bibitem[Chi et~al.(2022)Chi, Dong, Huang, Dai, Ma, Patra, Singhal, Bajaj,
  Song, and Wei]{chi2022representation}
Zewen Chi, Li~Dong, Shaohan Huang, Damai Dai, Shuming Ma, Barun Patra, Saksham
  Singhal, Payal Bajaj, Xia Song, and Furu Wei.
\newblock On the representation collapse of sparse mixture of experts.
\newblock \emph{arXiv preprint arXiv:2204.09179}, 2022.

\bibitem[Dagan et~al.(2005)Dagan, Glickman, and Magnini]{RTE1}
Ido Dagan, Oren Glickman, and Bernardo Magnini.
\newblock The {PASCAL} recognising textual entailment challenge.
\newblock In \emph{{MLCW}}, volume 3944 of \emph{Lecture Notes in Computer
  Science}, pages 177--190. Springer, 2005.

\bibitem[Dai et~al.(2022)Dai, Dong, Ma, Zheng, Sui, Chang, and
  Wei]{dai2022stablemoe}
Damai Dai, Li~Dong, Shuming Ma, Bo~Zheng, Zhifang Sui, Baobao Chang, and Furu
  Wei.
\newblock Stablemoe: Stable routing strategy for mixture of experts.
\newblock \emph{arXiv preprint arXiv:2204.08396}, 2022.

\bibitem[Devlin et~al.(2019)Devlin, Chang, Lee, and Toutanova]{BERT}
Jacob Devlin, Ming{-}Wei Chang, Kenton Lee, and Kristina Toutanova.
\newblock {BERT:} pre-training of deep bidirectional transformers for language
  understanding.
\newblock In \emph{{NAACL-HLT} {(1)}}, pages 4171--4186. Association for
  Computational Linguistics, 2019.

\bibitem[Dolan and Brockett(2005)]{MRPC}
William~B. Dolan and Chris Brockett.
\newblock Automatically constructing a corpus of sentential paraphrases.
\newblock In \emph{Proceedings of the Third International Workshop on
  Paraphrasing, IWP@IJCNLP 2005, Jeju Island, Korea, October 2005, 2005}. Asian
  Federation of Natural Language Processing, 2005.
\newblock URL \url{https://aclanthology.org/I05-5002/}.

\bibitem[Dosovitskiy et~al.(2020)Dosovitskiy, Beyer, Kolesnikov, Weissenborn,
  Zhai, Unterthiner, Dehghani, Minderer, Heigold, Gelly,
  et~al.]{dosovitskiy2020image}
Alexey Dosovitskiy, Lucas Beyer, Alexander Kolesnikov, Dirk Weissenborn,
  Xiaohua Zhai, Thomas Unterthiner, Mostafa Dehghani, Matthias Minderer, Georg
  Heigold, Sylvain Gelly, et~al.
\newblock An image is worth 16x16 words: Transformers for image recognition at
  scale.
\newblock \emph{arXiv preprint arXiv:2010.11929}, 2020.

\bibitem[Dua et~al.(2021)Dua, Bhosale, Goswami, Cross, Lewis, and
  Fan]{dua_tricks_2021}
Dheeru Dua, Shruti Bhosale, Vedanuj Goswami, James Cross, Mike Lewis, and
  Angela Fan.
\newblock Tricks for {Training} {Sparse} {Translation} {Models}.
\newblock 2021.
\newblock URL \url{http://www.statmt.org/wmt15/translation-task.html}.
\newblock arXiv: 2110.08246v1 ISBN: 2110.08246v1.

\bibitem[Fedus et~al.(2021{\natexlab{a}})Fedus, Brain, Zoph, and
  Shazeer]{fedus_switch_2021}
William Fedus, Google Brain, Barret Zoph, and Noam Shazeer.
\newblock {SWITCH} {TRANSFORMERS}: {SCALING} {TO} {TRILLION} {PARAMETER}
  {MODELS} {WITH} {SIMPLE} {AND} {EFFICIENT} {SPARSITY}.
\newblock 2021{\natexlab{a}}.
\newblock arXiv: 2101.03961v1.

\bibitem[Fedus et~al.(2021{\natexlab{b}})Fedus, Zoph, and
  Shazeer]{switch_transformer}
William Fedus, Barret Zoph, and Noam Shazeer.
\newblock Switch transformers: Scaling to trillion parameter models with simple
  and efficient sparsity.
\newblock \emph{CoRR}, abs/2101.03961, 2021{\natexlab{b}}.
\newblock URL \url{https://arxiv.org/abs/2101.03961}.

\bibitem[Foundation()]{wikidump}
Wikimedia Foundation.
\newblock Wikimedia downloads.
\newblock URL \url{https://dumps.wikimedia.org}.

\bibitem[Giampiccolo et~al.(2007)Giampiccolo, Magnini, Dagan, and Dolan]{RTE3}
Danilo Giampiccolo, Bernardo Magnini, Ido Dagan, and Bill Dolan.
\newblock The third {PASCAL} recognizing textual entailment challenge.
\newblock In \emph{ACL-PASCAL@ACL}, pages 1--9. Association for Computational
  Linguistics, 2007.

\bibitem[Haim et~al.(2006)Haim, Dagan, Dolan, Ferro, Giampiccolo, Magnini, and
  Szpektor]{RTE2}
R~Bar Haim, Ido Dagan, Bill Dolan, Lisa Ferro, Danilo Giampiccolo, Bernardo
  Magnini, and Idan Szpektor.
\newblock The second pascal recognising textual entailment challenge.
\newblock In \emph{Proceedings of the Second PASCAL Challenges Workshop on
  Recognising Textual Entailment}, 2006.

\bibitem[Lepikhin et~al.(2020)Lepikhin, Lee, Xu, Chen, Firat, Huang, Krikun,
  Shazeer, and Chen]{Gshards}
Dmitry Lepikhin, HyoukJoong Lee, Yuanzhong Xu, Dehao Chen, Orhan Firat, Yanping
  Huang, Maxim Krikun, Noam Shazeer, and Zhifeng Chen.
\newblock Gshard: Scaling giant models with conditional computation and
  automatic sharding.
\newblock \emph{CoRR}, abs/2006.16668, 2020.
\newblock URL \url{https://arxiv.org/abs/2006.16668}.

\bibitem[Lewis et~al.()Lewis, Bhosale, Dettmers, Goyal, and
  Zettlemoyer]{lewis_base_nodate}
Mike Lewis, Shruti Bhosale, Tim Dettmers, Naman Goyal, and Luke Zettlemoyer.
\newblock {BASE} {Layers}: {Simplifying} {Training} of {Large}, {Sparse}
  {Models}.
\newblock URL \url{https://github.com/pytorch/fairseq/}.
\newblock arXiv: 2103.16716v1.

\bibitem[Nie et~al.(2021)Nie, Cao, Miao, Ma, Xue, Miao, Yang, Yang, and
  Cui]{nie_dense--sparse_2021}
Xiaonan Nie, Shijie Cao, Xupeng Miao, Lingxiao Ma, Jilong Xue, Youshan Miao,
  Zichao Yang, Zhi Yang, and Bin Cui.
\newblock Dense-to-{Sparse} {Gate} for {Mixture}-of-{Experts}.
\newblock Technical Report arXiv:2112.14397, arXiv, December 2021.
\newblock URL \url{http://arxiv.org/abs/2112.14397}.
\newblock arXiv:2112.14397 [cs] type: article.

\bibitem[Rajbhandari et~al.()Rajbhandari, Li, Yao, Zhang, Aminabadi, Awan,
  Rasley, and Microsoft]{rajbhandari_deepspeed-moe_nodate}
Samyam Rajbhandari, Conglong Li, Zhewei Yao, Minjia Zhang, Reza~Yazdani
  Aminabadi, Ahmad Awan, Jeff Rasley, and Yuxiong~He Microsoft.
\newblock {DeepSpeed}-{MoE}: {Advancing} {Mixture}-of-{Experts} {Inference} and
  {Training} to {Power} {Next}-{Generation} {AI} {Scale}.
\newblock URL \url{https://github.com/microsoft/DeepSpeed}.
\newblock arXiv: 2201.05596v1.

\bibitem[Rajpurkar et~al.(2016)Rajpurkar, Zhang, Lopyrev, and Liang]{SQUAD1.1}
Pranav Rajpurkar, Jian Zhang, Konstantin Lopyrev, and Percy Liang.
\newblock Squad: 100, 000+ questions for machine comprehension of text.
\newblock In \emph{{EMNLP}}, pages 2383--2392. The Association for
  Computational Linguistics, 2016.

\bibitem[Rajpurkar et~al.(2018)Rajpurkar, Jia, and Liang]{SQUAD2.0}
Pranav Rajpurkar, Robin Jia, and Percy Liang.
\newblock Know what you don't know: Unanswerable questions for squad.
\newblock In \emph{{ACL} {(2)}}, pages 784--789. Association for Computational
  Linguistics, 2018.

\bibitem[Riquelme et~al.(2021)Riquelme, Puigcerver, Mustafa, Neumann, Jenatton,
  Pinto, Keysers, and Houlsby]{visual_moe}
Carlos Riquelme, Joan Puigcerver, Basil Mustafa, Maxim Neumann, Rodolphe
  Jenatton, Andr{\'{e}}~Susano Pinto, Daniel Keysers, and Neil Houlsby.
\newblock Scaling vision with sparse mixture of experts.
\newblock In \emph{NeurIPS}, pages 8583--8595, 2021.

\bibitem[Roller et~al.()Roller, Sukhbaatar, Szlam, and
  Weston]{roller_hash_nodate}
Stephen Roller, Sainbayar Sukhbaatar, Arthur Szlam, and Jason Weston.
\newblock Hash {Layers} {For} {Large} {Sparse} {Models}.
\newblock arXiv: 2106.04426v3.

\bibitem[Shazeer et~al.(2017)Shazeer, Mirhoseini, Maziarz, Davis, Le, Hinton,
  and Dean]{shazeer2017outrageously}
Noam Shazeer, Azalia Mirhoseini, Krzysztof Maziarz, Andy Davis, Quoc Le,
  Geoffrey Hinton, and Jeff Dean.
\newblock Outrageously large neural networks: The sparsely-gated
  mixture-of-experts layer.
\newblock \emph{arXiv preprint arXiv:1701.06538}, 2017.

\bibitem[Socher et~al.(2013)Socher, Perelygin, Wu, Chuang, Manning, Ng, and
  Potts]{SST2}
Richard Socher, Alex Perelygin, Jean Wu, Jason Chuang, Christopher~D. Manning,
  Andrew~Y. Ng, and Christopher Potts.
\newblock Recursive deep models for semantic compositionality over a sentiment
  treebank.
\newblock In \emph{{EMNLP}}, pages 1631--1642. {ACL}, 2013.

\bibitem[Turc et~al.(2019)Turc, Chang, Lee, and Toutanova]{turc2019well}
Iulia Turc, Ming-Wei Chang, Kenton Lee, and Kristina Toutanova.
\newblock Well-read students learn better: On the importance of pre-training
  compact models.
\newblock \emph{arXiv preprint arXiv:1908.08962}, 2019.

\bibitem[Wang et~al.(2019)Wang, Singh, Michael, Hill, Levy, and Bowman]{GLUE}
Alex Wang, Amanpreet Singh, Julian Michael, Felix Hill, Omer Levy, and
  Samuel~R. Bowman.
\newblock {GLUE:} {A} multi-task benchmark and analysis platform for natural
  language understanding.
\newblock In \emph{7th International Conference on Learning Representations,
  {ICLR} 2019, New Orleans, LA, USA, May 6-9, 2019}. OpenReview.net, 2019.
\newblock URL \url{https://openreview.net/forum?id=rJ4km2R5t7}.

\bibitem[Warstadt et~al.(2019)Warstadt, Singh, and Bowman]{CoLA}
Alex Warstadt, Amanpreet Singh, and Samuel~R. Bowman.
\newblock Neural network acceptability judgments.
\newblock \emph{Trans. Assoc. Comput. Linguistics}, 7:\penalty0 625--641, 2019.

\bibitem[Williams et~al.(2018)Williams, Nangia, and Bowman]{MNLI}
Adina Williams, Nikita Nangia, and Samuel~R. Bowman.
\newblock A broad-coverage challenge corpus for sentence understanding through
  inference.
\newblock In \emph{{NAACL-HLT}}, pages 1112--1122. Association for
  Computational Linguistics, 2018.

\bibitem[Xue et~al.()Xue, He, Ren, Lou, and You]{xue_one_nodate}
Fuzhao Xue, Xiaoxin He, Xiaozhe Ren, Yuxuan Lou, and Yang You.
\newblock One {Student} {Knows} {All} {Experts} {Know}: {From} {Sparse} to
  {Dense}.
\newblock arXiv: 2201.10890v2.

\bibitem[Zhu et~al.(2015)Zhu, Kiros, Zemel, Salakhutdinov, Urtasun, Torralba,
  and Fidler]{Zhu_2015_ICCV}
Yukun Zhu, Ryan Kiros, Rich Zemel, Ruslan Salakhutdinov, Raquel Urtasun,
  Antonio Torralba, and Sanja Fidler.
\newblock Aligning books and movies: Towards story-like visual explanations by
  watching movies and reading books.
\newblock In \emph{The IEEE International Conference on Computer Vision
  (ICCV)}, December 2015.

\end{thebibliography}

\newpage

\appendix

\section{Hyperparameters for Pre-training}
\label{app:params:pt}
Table~\ref{table:pthparam} presents the hyperparameters for pre-training. Table~\ref{table:mhparam} presents the hyperparameters of our MoE model.

\begin{table}[h]
\centering
\small
\begin{tabular}{lr}
\toprule
Hyperparameters & Value \\ \midrule
Optimizer & Adam \\
Training steps & 125,000 \\
Batch size & 2,048 \\
Adam $\epsilon$ & 1e-6 \\
Adam $\beta$ & (0.9, 0.98) \\
Maximum learning rate & 5e-4 \\
Learning rate schedule & Linear decay \\
Warmup steps & 10,000 \\
Weight decay & 0.01 \\
Transformer dropout & 0.1 \\
MoE balance loss weight & 1e-2 \\
\bottomrule
\end{tabular}
\caption{Pre-training hyperparameters}
\label{table:pthparam}
\end{table}

\begin{table}[h]
\centering
\small
\begin{tabular}{lr}
\toprule
Hyperparameters & Value \\ \midrule
Transformer blocks & 12 \\
Hidden size & 768 \\
FFN inner hidden size & 3,072 \\
Attention heads & 12 \\
Number of experts & 32 \\
\bottomrule
\end{tabular}
\caption{Model hyperparameters}
\label{table:mhparam}
\end{table}

\section{Hyperparameters for Fine-tuning}
\label{app:params:ft}
Table~\ref{table:fthparam} presents the hyperparameters for fine-tuning.

\begin{table}[h]
\centering
\renewcommand\tabcolsep{3.5pt}
\scalebox{0.73}{
\begin{tabular}{lrrrrrrrrr}
    \toprule
    \textbf{Hyperparamters} & \textbf{CoLA} & \textbf{STS-B} & \textbf{RTE} & \textbf{MNLI} & \textbf{SST-2} & \textbf{QQP} & \textbf{QNLI} & \textbf{MRPC} & \textbf{SQuAD} \\
    \midrule
    Batch Size & [32,16] & [32,16] & [32,16] & [32]  & [32]  & [32]  & [32]  & [32,16] & [32] \\
    Seed  & [1,2,3] & [1,2,3] & [1,2,3] & [2,3,5] & [2,3,5] & [2,3,5] & [2,3,5] & [1,2,3] & [1,2,3] \\
    Learning rate & [2,3,4,5]e-5 & [2,3,4,5]e-5 & [2,3,4,5]e-5 & [1,2,3,4]e-5 & [1,2,3,4]e-5 & [1,2,3,4]e-5 & [1,2,3,4]e-5 & [2,3,4,5]e-5 & [2,3,4]e-5 \\
    Warm up & [16, 10] & [16, 10] & [16, 10] & [16]  & [16]  & [16]  & [16]  & [16, 10] & [10] \\
    Epochs & [2,3,5,10] & [2,3,5,10] & [2,3,5,10] & [2,3,5] & [2,3,5] & [2,3,5] & [2,3,5] & [2,3,5,10] & [3] \\
    \bottomrule
    \end{tabular}
}
\caption{Hyperparameters for fine-tuning downstream tasks.}
\label{table:fthparam}
\end{table}

\section{More Results}

\begin{table}[htbp]
  \centering
  \caption{The complete GLUE results.}
  \resizebox{\linewidth}{!}{
    \begin{tabular}{ccccccccccc}
    \toprule
    \textbf{Settings} & \textbf{CoLA} & \textbf{STS-B} & \textbf{RTE} & \textbf{MNLI-m} & \textbf{MNLI-mm} & \textbf{SST-2} & \textbf{QQP} & \textbf{QNLI} & \textbf{MRPC} & \textbf{AVG} \\
    \midrule
    Dense-ft & 57.23  & 89.18  & 67.27  & 85.87  & 85.87  & 92.87  & 91.20  & 92.23  & 88.07  & 83.31  \\
    MoE-ft & 63.75  & 89.37  & 61.83  & 86.93  & 86.93  & 94.03  & 91.37  & 92.90  & 89.13  & 84.03  \\
    \midrule
    Two-pass-staged-drop & 58.32  & 88.64  & 60.32  & 86.30  & 85.90  & 90.00  & 90.50  & 91.40  & 86.90  & 82.03  \\
    Two-pass-eager-drop & 60.89  & 89.02  & 60.06  & 86.10  & 86.20  & 92.60  & 90.60  & 91.50  & 87.72  & 82.74  \\
    \midrule
    MoE-staged-pruning & 56.20  & 89.00  & \textbf{64.53} & 85.33  & 85.33  & 92.33  & 90.07  & 91.97  & 86.83  & 82.40  \\
    MoE-eager-pruning & \textbf{61.47} & \textbf{89.43} & 62.70  & \textbf{86.10} & \textbf{86.10} & \textbf{93.17} & \textbf{91.20} & \textbf{92.53} & \textbf{89.30} & \textbf{83.56} \\
    \bottomrule
    \end{tabular}}
  \label{tab:more_results}%
\end{table}




\end{document}